\documentclass{acmsiggraph}

\usepackage{amsmath}
\usepackage{graphicx}

\title{A Feature based Approach for Video Compression}

\author{Rajer Sindhu}
\pdfauthor{Rajer Sindhu}


\setcopyright{rightsretained}

\copyrightyear{2016}

\begin{document}

\maketitle

\begin{abstract}It is a high cost problem for panoramic image stitching via image matching algorithm and not practical for real-time performance. In this paper, we take full advantage ofHarris corner invariant characterization method light intensity parallel meaning, translation and rotation, and made a realtime panoramic image stitching algorithm. According to the basic characteristics and performance FPGA classical algorithm, several modules such as the feature point extraction, and matching description is to optimize the feature-based logic. Real-time optimization system to achieve high precision match. The new algorithm process the image from pixel domain and obtained from CCD camera Xilinx Spartan-6 hardware platform. After the image stitching algorithm, will eventually form a portable  interface to output high-definition content on the display. The results showed that, the proposed algorithm has higher precision with good real-time performance and robustness.

\end{abstract}

\keywordlist

\conceptlist

\printcopyright

\section{Introduction}
At present, the static image stitching algorithm has matured, but there is little research and application of video mosaics. The key is to match the real-time video splicing. But thanks to the matching method based on gray matching, greatly affected the accuracy of image matching, greatly affected the image of the little information poor and severely deformed gray areas of the image, as well as the region. Lowe made use SIFT feature matching descriptor \cite{1}\cite{2}. In contrast, light intensity and rotation changes and other conditions, can still accurate image feature extraction. The image matching methods in a wide range of applications, but the computational complexity, consume considerable resources. Harris corner than SIFT feature points calculated meaning easier and reduce the consumption of resources \cite{3}\cite{4}. However, due to limited software run faster, during the matching feature points, almost need a second feature point matching, can not meet the real-time requirements \cite{5}. Several modules of SIFT optimization algorithm attempts to achieve through SURF matching images \cite{6}, but he only introduced the main directions of the FPGA logic implementation, rather than the entire matching process. The present study showed, conducted based on SIFT algorithm \cite{7} SURF implementation, but because of the high complexity of describing the operation of the soft-core FPGA module is carried out \cite{8}\cite{9}. Due to the limited speed soft heart, the system does not achieve real-time real-time.

In order to meet the requirements of real-time display system, many designers prefer to use FPGA think parallel structures, a large number of logic array and fast signal processing. Harris corner detection, with good parallelism, you can rotate or scale variation FPGA real-time detection and image accurately extract the corner of \cite{10}. Point feature description method for screening algorithms can provide highly accurate information on the feature point matching, improve the matching accuracy. In this paper, Harris corner detection feature points are extracted and introduced the point description method in Xilinx Spartan-6 point match hardware platforms. The algorithm is the same for its performance and accuracy of the hardware for a reasonable optimized to achieve real-time, high precision and high stability of panoramic video stitching.

\begin{figure*}
    \includegraphics[width=18cm]{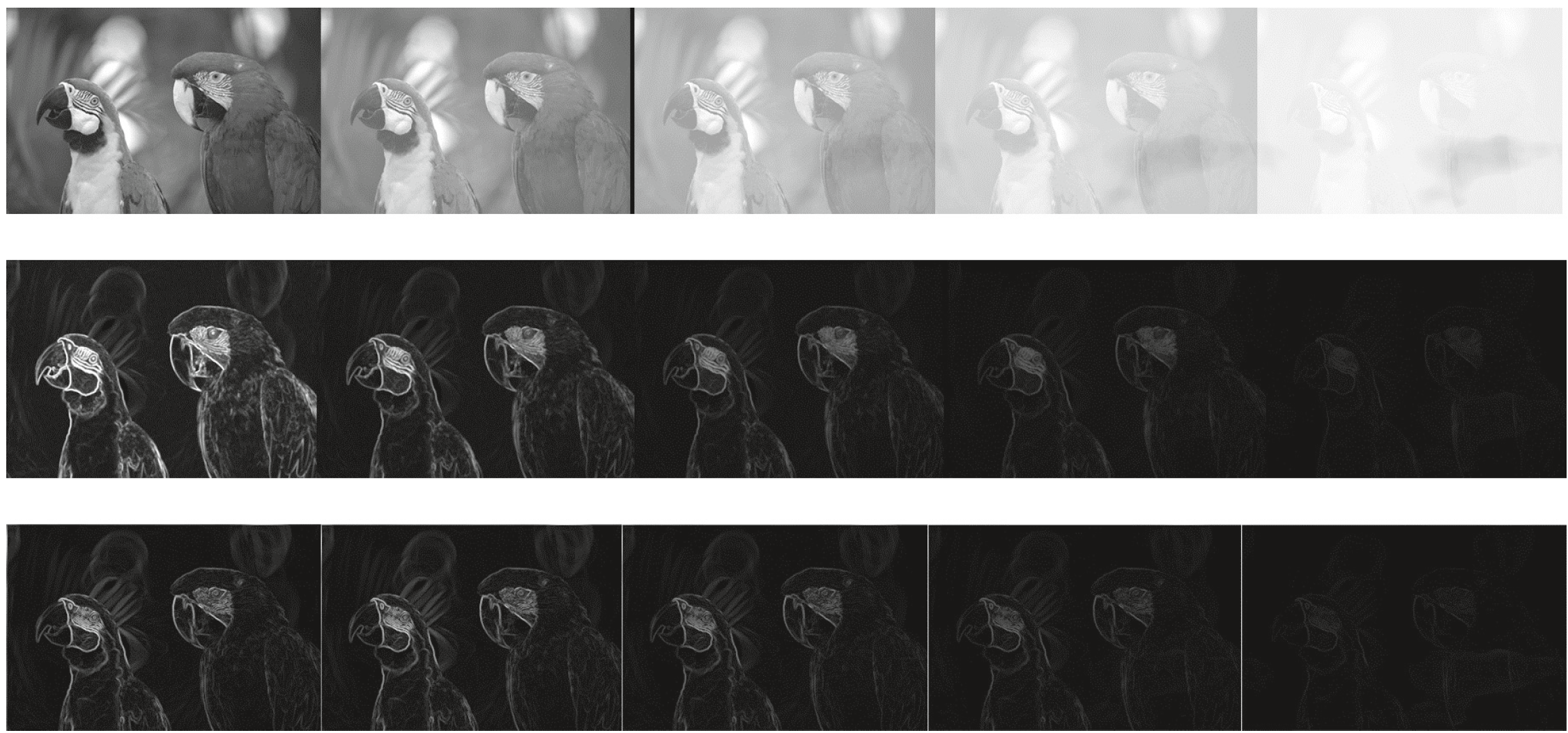}
    \caption{\label{fig 1:} Feature based video compression}
\end{figure*}

\section{Method Overview}
Assuming horizontal camera rotation speed is less than 30 degrees, a frame rate of 25 frames per second, then the frames will be less than 1.2 degree angular transformation. A rough estimate of the level of the video image of the moving distance of less than 10 percent of the image width. In hardware conditions, when the size of the overlapping area of the image size is 30 percent, received the highest match accuracy. Therefore, this article only half of the match image area. The match mode compared to full match match mode, it can improve the matching accuracy because feature points are concentrated, can effectively improve processing speed and conserve resources.

Screening algorithm can extract feature points from the image. In contrast, Harris operator to extract highly significant and stable curve calculation is very simple. When Harris corner does not perform very well, changes in scale, but they still choose notable feature points of the image as little effect on the scale of video splicing matching process. Then the feature vector described method for describing the calculation. Time proportional to the number of feature points of features described, it is necessary to ensure real-time algorithms, in order to ensure real-time algorithm. In the non-maxima suppression, the use of non-maximal inhibition threshold, adaptive feature point extraction. The threshold value calculation formula is as follows:

\subsection{Feature Extraction}
Classic SIFT feature described in the Gaussian cone image. Since Gauss cone is not applied in the new algorithm, it is necessary to optimize the classical methods. Specific procedures are as follows: (1) the main direction of extraction. Lowe's SIFT Gaussian filtered images are calculated to obtain the modulus value M (x, y) of the gradient of Gaussian distribution by processing, wherein the parameters ? satisfies the equation (1), and the radius of the window is 3 × 1.5. However, the algorithm of the histograms three three regions feature points near the original image is not necessary to do the Gaussian distribution processing module, thereby greatly reducing the amount of calculation. (2) calculate the area of the descriptor. the neighborhood features size 4 × 4 points around the region, and each region is equal to the area used to calculate the main direction of the corner, which is 3 × 3. in bilinear interpolation and axis of rotation is considered, should be set to calculate the area of a half. Where (×) is the gradient magnitude pixels. Rows and columns adjacent seed points and DC Dr. represent contributions pixels. Make contributions on behalf of the direction of the pixel to its neighbors. Potassium, rice, nitrogen are 0 and 1.

\begin{figure}
    \includegraphics[width=7cm]{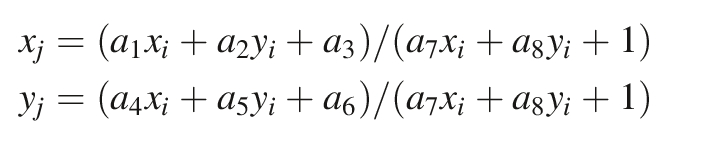}
\end{figure}

For real-time video stitching algorithm proposed FPGA-based hardware logic implementation, which mainly consists of three parts, namely, adaptive Harris corner detection, characterization and feature matching. Specific procedures shown in Fig 2. The module leverages the FPGA design can simultaneously pipeline and store images using Harris corner detection complete video stream each image feature point extraction, and then the coordinate values of the feature points.

\subsection{Video Compression}
In order to make the angle it has different characteristics, which is useful for image matching, gradation characteristic angle and surrounding area herein described quantitatively generated angular descriptors, which is the key and difficult in hardware implementations. As shown in Figure 5, the part extracted by the image gradient magnitude and direction of the main direction and description generation computing. Calculated gradient magnitude and direction. In order to ensure real-time, when extracting Harris corner point, we use the results of the X and Y directions of the derivative to calculate the magnitude and direction of the gradient of image pixels \cite{11}.

To achieve root, and trigonometric functions in the  is not easy, it uses  algorithm operation, saving the magnitude and direction of the results for future use. In computing, it is the implementation of these complex nonlinear function of the hardware implementation is very convenient, the algorithm uses only shift and add in. To improve the stability of the system, the paper called directly embedded core from  IP, do the calculation.  Extraction main direction angles. Reads the amplitude and direction of the 3 × 3 area of RAM from the corner, and with $\theta \times \pi$ to judge angles are integers \cite{12}. Add the same gradient direction histogram integral corresponding amplitudes obtained. Then compare the histogram of all the columns and finally get the main direction angle $\theta$. Angle range calculated, the time interval, 36, will be reduced to 18 interval map directions, reducing the time and resource consumption \cite{13}.

In order to compare our approach with other image segmentation algorithm, the average value of  measures as a performance of the method. We compare our method has the following methods. Comparative results are summarized in Table 2, which shows that our proposed method is better than most other methods. figure 2. Three image segmentation method under different lighting conditions of application examples. (A) The original color of retinal images. (B) segmentation results with our approach. A first hand expert (online color image) by the segmentation results. Based on 15 points retinal vascular image. Described angle characteristics. Many described the system as a whole DSP or FPGA embedded software constructs avoid counting capacity, which will increase costs and reduce the computing speed. In this paper, cleverly calculated, MATLAB image is irrelevant, and the results stored in a lookup table in the form to register. This article simply reads the values in the table to complete the description \cite{14}.
\begin{figure}
    \includegraphics[width=8cm]{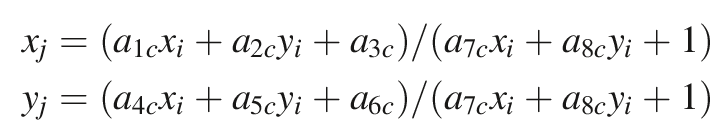}
\end{figure}

\section{Experiments}
Take indoor scenes, for example, the use of MATLAB with the classical Harris matching algorithm two consecutive static image stitching process simulation, classic SIFT matching algorithm, the new algorithm in this document and the adaptation of the hardware platform as shown in Table 1 and 10, respectively, shown. Table 1 shows the images obtained by using different algorithms in FIG. 10 (1) mosaic image. This article can be seen from the data, compared with the classical algorithm, the new algorithm can extract a feature point having a higher significance and has a higher matching accuracy. The applicable version of the hardware to maintain good matching effect without reducing the matching accuracy.

\begin{figure}
    \includegraphics[width=8cm]{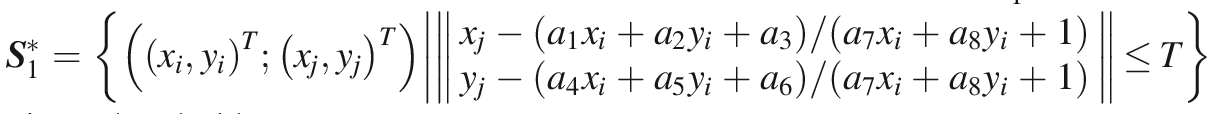}
\end{figure}
\begin{figure*}
    \includegraphics[width=18cm]{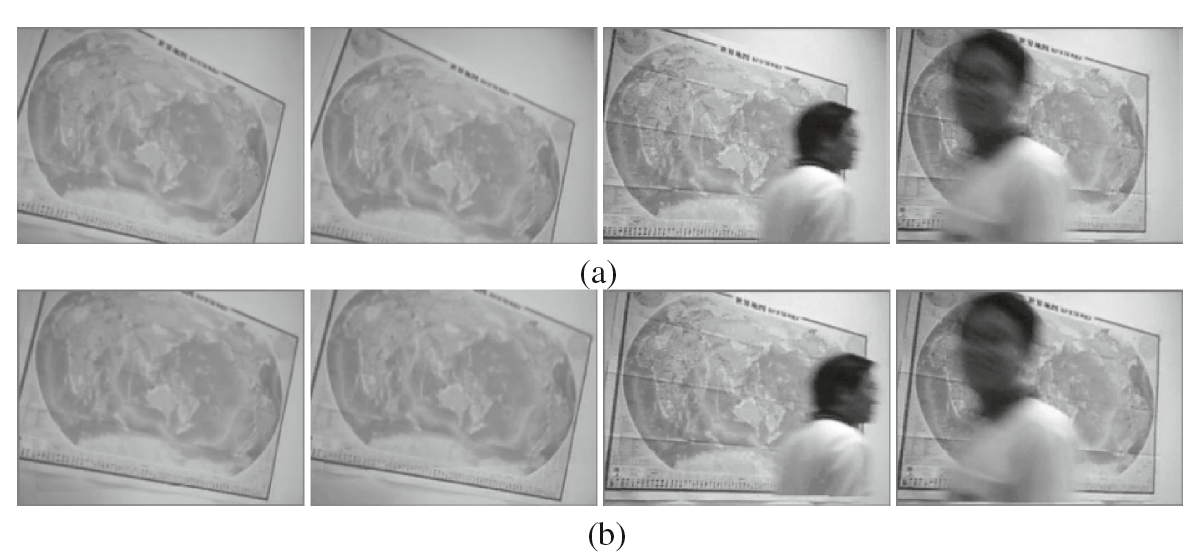}
    \caption{\label{fig 1:} The result of video frame before/after compression}
\end{figure*}

As shown in Table, the algorithm consumes less logic resources in Figure 3, but a large number of on-chip memory resources. This is because the implementation of the most complex operations in the form of a lookup table stored in on-chip memory, which reduces logic resources consumption, improve the operating speed of the algorithm. At the same time, the storage and display of video image data stored in the register DDR3 chip RAM reducing the consumption of resources on the chip.
\begin{figure}
    \includegraphics[width=9cm]{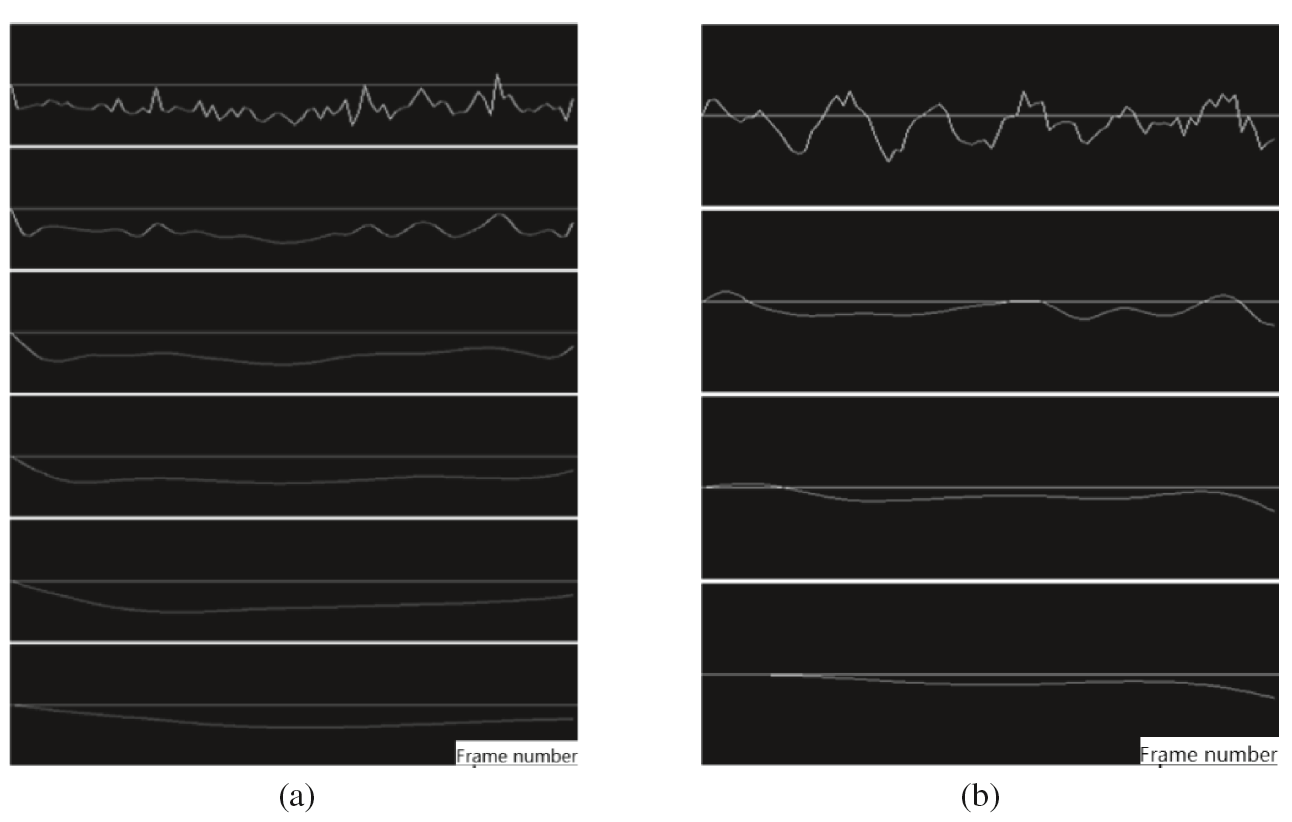}
    \caption{\label{fig 1:} Motion filtering results}
\end{figure}

\begin{figure}
    \includegraphics[width=9cm]{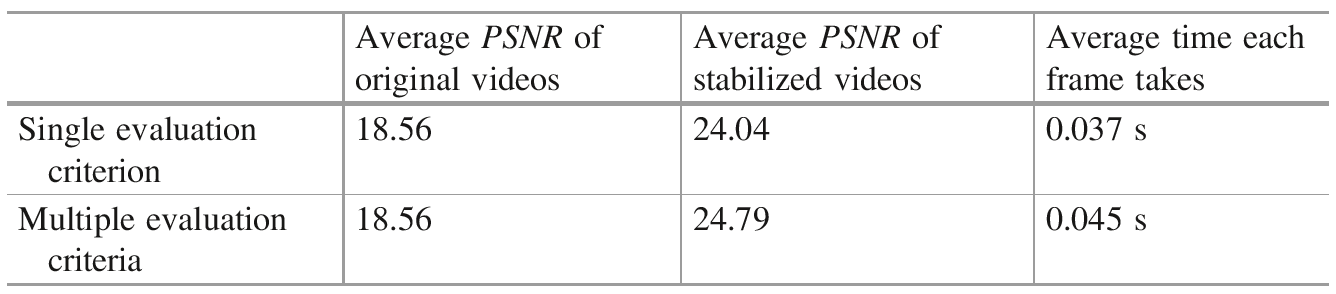}
    \caption{\label{fig 1:} Assessment of several video stabilization algorithm}
\end{figure}

\begin{figure}
    \includegraphics[width=9cm]{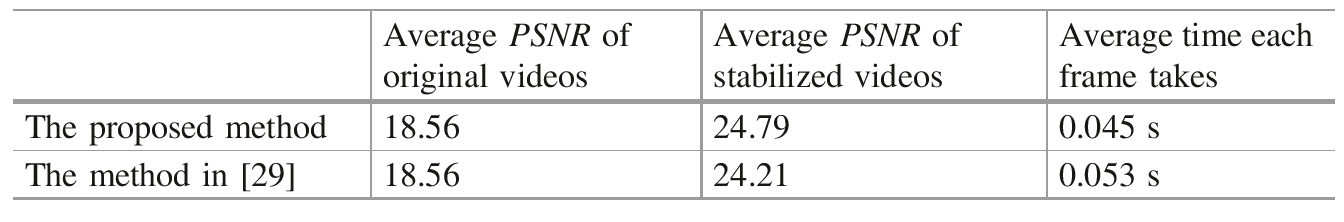}
    \caption{\label{fig 1:} Questionnaire and average scores assessment results}
\end{figure}

\section{Conclusion}
This paper discusses the image stitching algorithm in FPGA implementation process, the full use of the invariant characterization methods in the light intensity changes SIFT, translation and rotation, and the properties of the image matching based on Harris corner points. Depending on the hardware features of the new algorithm is optimized. The image display system showed FPGA algorithm has good real-time and meet the high precision image sequence matching the needs of application performance test results of the video image. However, due to hardware limitations, how to achieve high-precision matching conditions, such as an image scale changes or severe image distortion, needs further study.

\bibliographystyle{acmsiggraph}
\nocite{*}
\bibliography{template}
\end{document}